\documentclass[preprint,12pt]{elsarticle}

\usepackage{amssymb}
\usepackage{xcolor}
\usepackage{soul}
\usepackage{multirow}
\usepackage{subcaption}
\usepackage{caption}
\usepackage{graphicx}
\usepackage{booktabs}
\usepackage{adjustbox}

\journal{Smart Health}

\begin{document}

\begin{frontmatter}



\title{Language-Assisted Deep Learning for Autistic Behaviors Recognition}


\author[inst1]{Andong Deng}
\author[inst1]{Taojiannan Yang}
\author[inst1]{Chen Chen} 
\affiliation[inst1]{organization={Center for Research in Computer Vision, University of Central Florida},
            city={Orlando},
            state={FL},
            country={USA}}

\author[inst2]{Qian Chen}
\affiliation[inst2]{organization={Dept. of Electrical and Computer Engineering, University of Texas at San Antonio},
            city={San Antonio},
            state={TX},
            country={USA}}

\author[inst3]{Leslie Neely}
\affiliation[inst3]{organization={Child and Adolescent Policy and Research Institute, University of Texas at San Antonio},
            city={San Antonio},
            state={TX},
            country={USA}}
\author[inst4]{Sakiko Oyama}
\affiliation[inst4]{organization={Dept. of Kinesiology, University of Texas at San Antonio},
            city={San Antonio},
            state={TX},
            country={USA}}

\begin{abstract}
  
  Correctly recognizing the behaviors of children with Autism Spectrum Disorder (ASD) is of vital importance for the diagnosis of Autism and timely early intervention. However, the observation and recording during the treatment from the parents of autistic children may not be accurate and objective. In such cases, automatic recognition systems based on computer vision and machine learning (in particular deep learning) technology can alleviate this issue to a large extent. Existing human action recognition models can now achieve impressive performance on challenging activity datasets, e.g., daily activity, and sports activity. However, problem behaviors in children with ASD are very different from these general activities, and recognizing these problem behaviors via computer vision is less studied. 
  In this paper, we first evaluate a strong baseline for action recognition, i.e., Video Swin Transformer, on two autism behaviors datasets (SSBD and ESBD) and show that it can achieve high accuracy and outperform the previous methods by a large margin, demonstrating the feasibility of vision-based problem behaviors recognition. Moreover, we propose language-assisted training to further enhance the action recognition performance. Specifically, we develop a two-branch multimodal deep learning framework by incorporating the "freely available" language description for each type of problem behavior. Experimental results demonstrate that incorporating additional language supervision can bring an obvious performance boost for the autism problem behaviors recognition task as compared to using the video information only (i.e., 3.49$\%$ improvement on ESBD and 1.46$\%$ on SSBD). \textit{Project website: https://dengandong.github.io/UCF-ASD}
  
\end{abstract}



\begin{keyword}
Autistic Behaviors \sep Deep Learning 
\PACS 0000 \sep 1111
\MSC 0000 \sep 1111
\end{keyword}

\end{frontmatter}


\section{Introduction}
Autism has been an increasing health problem for infants and toddlers~\cite{rutter1968concepts,lord2006autism,hertz2009rise}. According to estimates from CDC's Autism and Developmental Disabilities Monitoring (ADDM) Network in 2020, about 1 in 44 U.S. children has been identified with autism spectrum disorder (ASD)\cite{christensen2018prevalence} \footnote{https://www.cdc.gov/ncbddd/autism/data.html}. 
Autistic children usually have disorders in social interactions, and communications and may have restricted and repetitive behaviors, which can be generally termed Autism Spectrum Disorder (ASD). ASD could bring lifelong disability if patients were not diagnosed and/or treated at an early stage. 
However, recent studies~\cite{lord2006autism} reveal that many children do not receive a final diagnosis until they grow much older (i.e., after 3 years old). The discovery and diagnosis of ASD at an early age (i.e., before 2 years old) is still a challenging problem.

Discriminating the problem behaviors of autistic children (e.g., self-injury) is a promising method for early diagnosis since such restricted or repetitive actions show some particular patterns which can be distinguished from other normal ones. Clinic physicians have defined several categories of such behavior disorders~\cite{lam2007repetitive} including  stereotypic behavior, self-injurious behaviors, ritualistic/sameness behavior, compulsive behavior, and restricted behavior. Stereotypic behavior is defined as behavior that is for no purpose but is performed repetitively. Self-injurious behaviors include ruthless actions that may bring potential danger to the patients themselves. Ritualistic or sameness behaviors are those similar actions in daily life. Compulsive behaviors are also some repeated actions like the stereotypic type but with clear rules. Restricted behaviors mean patients usually have limited attention to target objects. In common clinical cases, physicians can observe these abnormalities that children may occasionally flap their arms and repetitively hit their heads with their hands or on other solid objects during the time with their parents. Ideally, parents are the best choice to record such symptoms; however, due to possible emotional bias, their assessment can be far from objective and accurate.

Recently, the computer vision community has witnessed rapid development in the recognition models in both images~\cite{he2016deep, krizhevsky2017imagenet}  and videos~\cite{carreira2017quo, liu2022video}. Human action recognition models have been validated to achieve promising performance using various datasets,  and have been deployed in many practical application scenarios including video abnormal detection~\cite{Sultani_2018_CVPR, chen2021automated, lv2021learning}. The previous successful experience inspires Artificial Intelligence (AI)-based ASD diagnosis methods to analyze problem behaviors on two public ASD datasets, namely,  ESBD~\cite{negin2021vision} (Expanded Stereotype Behavior Dataset) and SSBD~\cite{rajagopalan2013self} (Self-Stimulatory Behavior Dataset). Computer vision models have been developed to evaluate the problem behavior recognition performance based on these two datasets. Ali et al.~\cite{ali2022video} train I3D~\cite{carreira2017quo} on SSBD and obtain 76.92$\%$ accuracy. Nagin et al.~\cite{negin2021vision} propose to utilize 3DCNN~\cite{zhang2017learning} and ConvLSTM~\cite{shi2015convolutional} on ESBD and obtain $42\%$ and $74\%$ accuracy, respectively. 

Although these initial studies show that the vision-based solutions are applicable to problem behavior recognition in children with ASD, the recognition performance is not very high (less than $85\%$). We also find that previous studies ignore the textual description of the corresponding stimming behaviors, which can be as discriminative as videos and can provide supervision for the recognition task. Numerous studies in other research fields have demonstrated the effectiveness of the combination of human language and visual data in a unified model, since the correspondence between the two modalities can provide abundant supervision for the model to learn. Some vision-language models~\cite {radford2021learning} can recognize objects without using labeled data, which is a revolution for traditional vision-only models. In this paper, we leverage both videos and the language descriptions of each stimming behavior class to supervise the model training. Specifically, we utilize a pre-trained language model and our visual network to obtain both text features and visual features, respectively. Afterward,  we minimize the classification objective while maximizing the cosine similarity between language and visual features, by which we can transfer knowledge, i.e., the vision-language correlations, from the pre-trained language model to our visual features. Our experiments demonstrate that language description, which could serve as additional supervision, can significantly enhance AI-based autistic behavior recognition.

The main contributions of this work can be summarized as follows:
\begin{itemize}
    \item We adopt one of the best video action recognition models, namely \textit{Video Swin Transformer}~\cite{liu2022video}, and train it for autism action recognition tasks. The experimental results on two popular autism behaviors datasets (ESBD and SSBD) show that our trained model outperforms previous vision-based methods by a significant margin, leading to a strong baseline for autism action recognition in videos. 
    
    \item We propose to incorporate the textual description of each autism action class into the visual training model. By introducing such free supervision, we further improve the autism action recognition performance without collecting more labeled data. Since autism behavior videos could be hard to obtain/re-obtain due to privacy, time, and non-repeatability, this training paradigm could largely alleviate the problem of training data scarcity. 
    
    \item Although our multimodal framework uses extra language supervision for training the model, the computational cost would not be increased in the test/inference phase since we only use the visual branch for testing.
\end{itemize}
\section{Related Work}

\subsection{Vision-Based Recognition Models}
Computer vision is in a blossoming era with the fast development of deep learning. A variety of sophisticated neural networks have been trained with extremely deep layers to solve a wide range of real-world problems including image classification, object detection, action recognition, and localization. Videos have become a main-stream data format to record human activities, and they are widely shared on social media platforms. Action recognition, as one of the most important tasks in the area of video understanding, therefore has been attracting attention.  In the early period, researchers used 2D convolutional neural networks to extract the features of selected keyframes of a video stream, and train a classifier to recognize actions~\cite{karpathy2014large}. The early two-stream convolutional networks leverage both dense RGB frames and optical flows. This model can process temporal information because flow data is complementary to static frames. 
The lengths of videos are varied from  minutes to hours; thus, models trained with dense frames are also criticized for ignoring long-range temporal information. To address this issue, temporal segment networks~\cite{wang2016temporal} evenly partition the whole frame sequence into chunks, and randomly samples one frame from each chunk. Based on such methods, the frames fed into the network, though sparse, can cover a temporal event as completely as possible. However, 2D convolution  cannot process temporal correlations. 3D convolution networks were further proposed to solve this problem such as the inflated 3D networks (I3D)~\cite{carreira2017quo}. I3D could inflate 2D convolution kernels trained on large image classification datasets~\cite {deng2009imagenet} that have abundant visual information into 3D kernels. In this case, the training effort has been largely reduced.

Nevertheless, 3D convolution operations result in much higher computational complexity than their 2D counterparts. It is not practical to use 3D convolution networks to analyze complicated big data if there are not enough computing resources.
In recent years, a new architecture paradigm, Transformer~\cite{vaswani2017attention}, with stacked self-attention layers  as its basic operation block to model \textit{long-range correlations} in the data, outperforms previous works and has become the leading-edge AI technology in both vision and language. As one of the bests, Swin Transformer~\cite{liu2021swin}, built with a shifted window attention layer, has achieved state-of-the-art performance in all of the basic vision tasks, i.e.,  image classification (87.3$\%$ top-1 accuracy on ImageNet~\cite{deng2009imagenet}), object detection (58.1$\%$ AP on COCO~\cite{lin2014microsoft}) and semantic segmentation (62.8$\%$ mIoU on ADE20K~\cite{zhou2019semantic}). VideoSwin~\cite{liu2022video}, as the video version of Swin Transformer,  outperforms many cutting-edge video models such as TSN~\cite{wang2016temporal}, I3D~\cite{carreira2017quo} in action recognition tasks.  Thus, we choose VideoSwin for autism behavior recognition to expressively model the long-range correlations in the videos, and our experimental results also validate its outstanding performance and versatility.

\subsection{Vision-Language Recognition Models}
Natural language and vision are equally important in the AI community. Recently, investigating the intrinsic correspondence between vision and language has been a popular trend, which can reduce  data annotation burdens, and take advantage of language's abundant supervision.
Many strong vision-language models have been proposed based on pretraining on large-scale unlabeled datasets including DALL-E~\cite{ramesh2021zero}, Florence~\cite{yuan2021florence} and CLIP~\cite{radford2021learning}. The performance of direct evaluation (without training) on the target dataset using those strong vision-language models even surpasses the models well-trained with labeled datasets.  Most of these vision-language models are designed from contrastive learning that pulls the features from different modalities as close as possible in the feature space. Such design allows models to learn the cross-modal correlations much better than before. In the testing stage, visual outputs are matched to their most similar textual features to enhance their classification accuracy. 
Previous study~\cite{xiang2022language} has validated that the correlated language information provides strong support as additional supervision for vision tasks.  Based on the success of vision-language models, this work adopts a similar model and generates detailed descriptions for each class from two well-known Autism datasets ESBD and SSBD. Afterward, we obtain the textual features of the dataset by using the pretrained language encoder~\cite{radford2021learning}. For model training, we leverage an additional loss function to pull close the language features and visual features extracted by the language encoder and our visual network, respectively, which endows abundant vision-language knowledge from the pretrained model to our action recognition model. The final results of our model surpass previous studies significantly (Section \ref{sec:result}), which proves the effectiveness of our method.

\begin{figure}[t]
\centering
    \includegraphics[width=14cm]{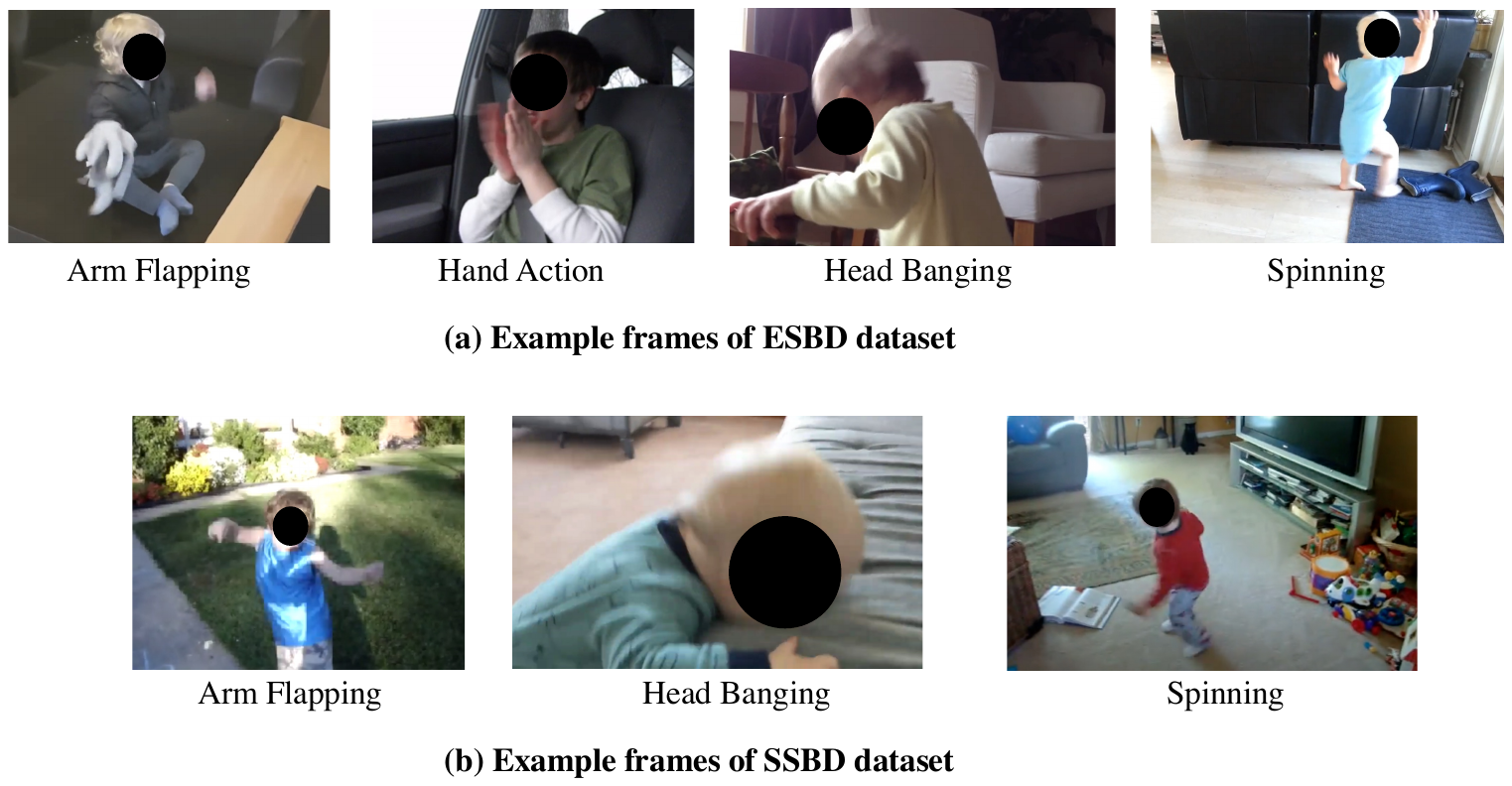}
    \vspace{-5em}
    \caption{Example video frames from the ESBD and SSBD datasets for ASD behavior recognition in children. }
    \vspace{-0.5em}
    \label{examples}
\end{figure}

\subsection{Vision-based ASD Behavior Analysis}
Computer vision, as one of the most important branches of AI, has proved its promising ability in massive real-world applications, specifically in the medical domain.  Advanced computer vision has been utilized for Computed Tomography (CT)~\cite{8355700}, Magnetic Resonance Imaging (MRI)~\cite{knoll2020deep}, biomedical image segmentation~\cite{ronneberger2015u} for many decades, and has achieved thrilling progress. Computer vision also facilitates the development of ASD diagnosis, since it helps to efficiently analyze stimming behaviors, which are one of the most important stages for autism final diagnosis. In 2013, Rajagopalan et al.~\cite{rajagopalan2013self} proposed a Self-Stimulatory Behaviour Dataset (SSBD) to investigate automatic behavior analysis in unlimited settings. This study utilizes  the traditional Space-Time Interest Points (STIP)~\cite{laptev2005space} combined with Bag of Words (BOW) to train a 3-class classifier. More recently, many deep learning methods have been used in stimming behavior analysis. The old SSBD dataset, therefore, is out of date because the dataset is small-scaled and some of the video URLs are inaccessible now.  Negin et al.~\cite{negin2021vision} contribute a new dataset namely ESBD, and train a 3D CNN and a ConvLSTM to evaluate model efficacy in autism diagnosis. According to their results, the deep learning methods they use are not better than traditional vision methods such as using HOF descriptors for feature extraction and MLP for classification. Wei et al.~\cite{wei2022vision} leverage Multi-Stage Temporal Convolutional Network and obtain 83$\%$ accuracy on SSBD. Neither of them considers incorporating the most state-of-the-art recognition models for ASD diagnosis and adopts visual modality models only. To enhance the ASD diagnosis and accurately recognize abnormal behaviors from these two datasets, we first train a strong baseline using Video Swin Transformer and then insert language data into the model for better recognition performance. The experimental results reveal that Video Swin Transformer outperforms previous methods considerably and the additional language supervision can further improve the recognition accuracy.

\begin{figure}[t]
\centering
    \includegraphics[width=14cm]{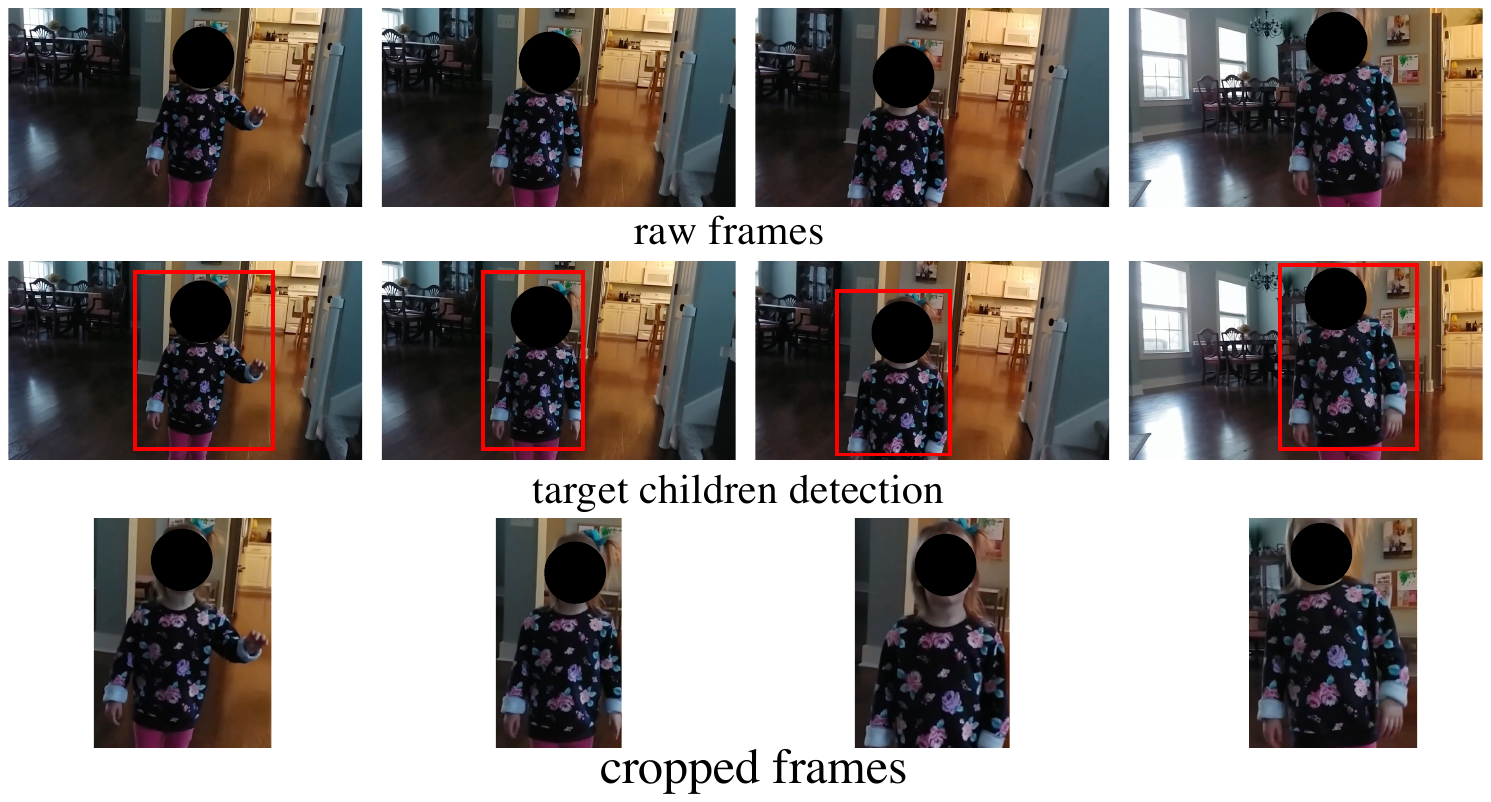}
    \vspace{-5em}
    \caption{Illustration of the data preprocessing steps. We first detect the target child in video frames via the YOLO v5 object detector and then crop the child region with the detected bounding boxes. These cropped frames with arbitrary sizes will be resized to the same size, i.e., $224\times 224$, before being fed into the action recognition network. }
    \label{preprocess}
\end{figure}
\section{Method}
\subsection{Datasets and Pre-processing}

There are two popular autism behavior datasets: Self-Stimulatory Behavior Dataset (SSBD)~\cite{rajagopalan2013self} and Expanded Stereotype Behavior Dataset (ESBD)~\cite{negin2021vision}. SSBD contains 75 videos of stimming actions of children with autism spectrum disorder which were published on public domain websites\footnote{https://rolandgoecke.net/research/datasets/ssbd/}, with an average duration of 90 seconds per video. These videos were recorded in uncontrolled natural settings by parents/caregivers. SSBD contains three typical action classes: Arm Flapping, Head Banging, and Spinning. Arm Flapping is a stimming behavior when autistic children cannot normally communicate with people around them or they have excessive energy. Head banging is typically a self-injurious action where children will use their own hands to hit on the head or just directly hit other solid objects (e.g., desks and walls) with their heads. Spinning is also a stimming behavior that the children often turn around their whole body repeatedly. Each category includes 25 video samples, but some video URLs are no longer available, and we only obtain 59 videos: 19 for Arm flapping and 20 for both Head banging and Spinning. The second dataset ESBD contains 99 YouTube videos, and the average video duration is about 2 minutes. Besides the three categories of SSBD, ESBD also includes the fourth category of Hand Action; thus ESBD contains  35 videos for Arm flapping, 13 videos for Hand action, 24 videos for Head banging, and 37 videos for Spinning. Note that hand action and arm flapping are different, hand action focuses more on the finger actions and can describe more actions than arm flapping. Similar to SSBD, due to poor maintenance, we only obtained 89 videos of ESBD, namely 29 videos for Arm flapping, 13 videos for Hand action, 16 videos for Head banging, and 31 videos for Spinning. Examples of these two datasets are presented in Figure ~\ref{examples}.

These two datasets are noisy and contain a large portion of the background or other subjects. To enhance recognition accuracy, we first preprocess the videos to obtain cleaner data that include target children performing ASD behaviors only. 
To this end, we leverage one of the most popular object detection models YOLOv5~\cite{bochkovskiy2020yolov4}, which is pretrained on the COCO~\cite{lin2014microsoft} dataset, as our target child detector (i.e., person detector).
With the help of YOLOv5, we obtain the cropped frames of the target children and behaviors. As presented in Figure~\ref{preprocess}, the cropped frame is much more focused on the child region than the raw data, which leads to a more stable training process and better action recognition results. In addition, we segment each video into several 30-frame clips to extract as much information as possible, and this video segmentation setting also increases the number of training samples. Our final training data, therefore, contains 1,437 and 1,030 clips for ESBD and SSBD, respectively.


\subsection{Vision-only Model}
\begin{figure*}[h]
\centering
    \includegraphics[width=12cm]{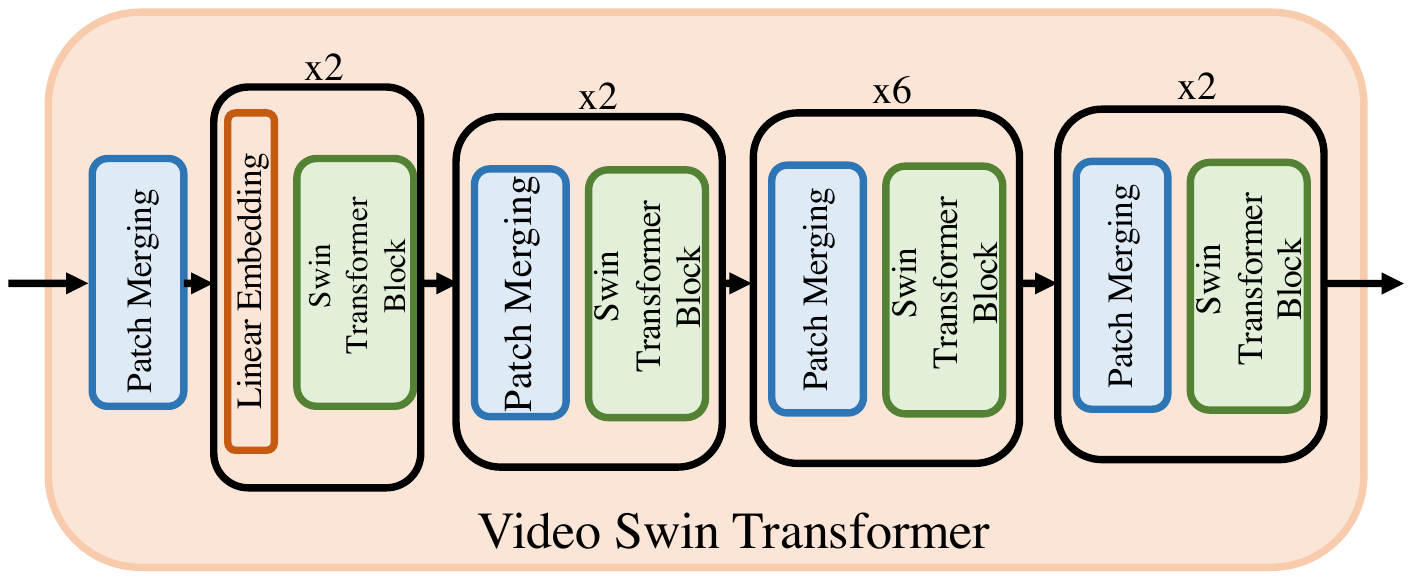}
    \vspace{-5em}
    \caption{Illustration of Video Swin Transformer with basic configuration.}
    \label{videoswin}
\end{figure*}

We first investigate the vision-only model for autism behavior recognition to set up a strong baseline. In previous studies~\cite{negin2021vision, wei2022vision, ali2022video}, several vision-based methods have been proposed and obtained promising performance on public autism behaviors datasets. However, the models they utilized are quite out-of-date and did not achieve satisfactory recognition results. More recent high-performing video action recognition models could benefit this task. Video Swin Transformer~\cite{liu2022video} is currently the state-of-the-art action recognition model which uses vision transformer architecture to explicitly model the long-term correlations in the video. 
We believe the long-range correlation modeling capability of the transformer model can also achieve high accuracy on autism behavior analysis even though there exists a domain gap between autism behavior datasets and regular action datasets. Basically, Video Swin Transformer contains 4 blocks, each of which is constructed with shifted window attention mechanism as follows:
\begin{equation}
Q = x \cdot W_Q,
\end{equation}
\begin{equation}
K = x \cdot W_K,
\end{equation}
\begin{equation}
V = x \cdot W_V,
\end{equation}
\begin{equation}
Attention(Q, K, V) = softmax(QK^T/\sqrt{d_k})V,
\end{equation}
where $x$ is the input feature of the attention layer, and $W_Q$, $W_K$, $W_V$ the projection weights. It should be noted that in shifted window attention the input features are segmented into chunks along the dimension axis via a predefined window, whose size is much smaller than the dimension size itself, before performing the attention operation. In this way, the computational complexity can be obviously reduced.
We follow the basic configurations [2, 2, 6, 2] (the number indicates the attention layers in each block), as shwon in Figure~\ref{videoswin},  which is the most popular one that is balanced between accuracy and efficiency. During training, we use the regular cross-entropy as the classification loss function. 
\vspace{-3em}

\begin{figure*}[t]
\centering
    \adjustbox{margin=-1in}{
    \includegraphics[width=16cm]{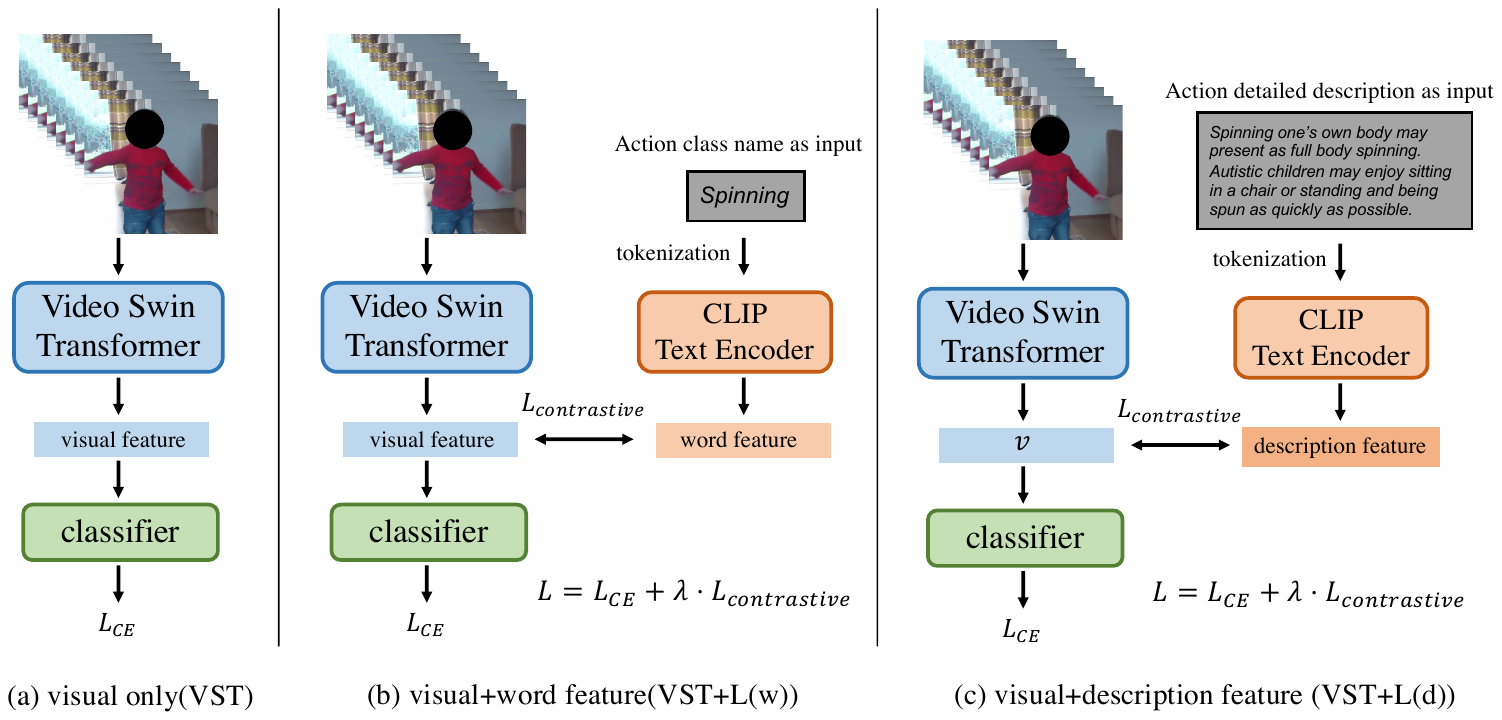}}
    \vspace{-5em}
    \caption{Comparison of the visual only Video Swin Transformer (VST) and the proposed language-assisted framework (VST+L). For VST+L, we also propose two alternatives, one of which utilizes the action class name as the language input, which is noted as VST+L(w), and another one, VST+L(d), leverages more detailed descriptions for each action class, leading to more informative language features. The comparison of the two alternatives can be found in Table~\ref{text input}. Generally, the input video is processed by the Video Swin Transformer network for visual feature extraction and the text description of the action in this video is processed by the CLIP text encoder for textural feature extraction. We use cross-entropy $L_{CE}$ as the classification loss and add the contrastive loss ($L_{constrastive}$) to enforce the paired visual and textural features are close to each other. It should be noted that language supervision is only used in the model training stage to enhance the visual feature representation. In the test/inference stage, only the visual branch is used to extract the feature from a test video for action recognition. }
    \vspace{-1em}
    \label{vl}
\end{figure*}

\subsection{Language-assisted Video Swin Transformer}

\begin{table}[h]\small%
\caption{Text descriptions of problem behaviors in children with ASD. These descriptions are generated from related publications, and we found the detailed descriptions could be better than the name of behavior categories. The comparison results are shown in Table~\ref{text input}.}
\centering
\scalebox{0.8}{
\setlength{\tabcolsep}{3mm}{
\begin{tabular}{l}
\toprule[1pt]
\textbf{Arm flapping}: Arm flapping is one of the stimming behaviors that involve \\
repetitive movement of the arms and hands. It's often used as a way to release \\
excess energy or stimulate the senses. \\  
\textbf{Head banging}: Head Banging is a self-injurious behavior for children with \\
autism spectrum disorder children who are sensitive to noise are often aggressive \\
and will hit their heads to distract themselves from the pain.\\
\textbf{Spinning}: Spinning one’s own body may present as full-body spinning. Autistic \\
children may enjoy sitting in a chair or standing and being spun as quickly as possible. \\  
\textbf{Hand action}: Finger flicking is a repeated movement involving fingers using an \\
almost “snapping” motion. The repetitive motion of finger flicking close to the face \\
lets the child knows where their body is in relation to space and other objects. \\  \bottomrule[1pt]
\end{tabular}}}
\vspace{-0.5em}
\label{description}
\end{table}

Vision-language models have been demonstrated to have extraordinary ability of representation and can fit a variety of vision scenarios since the intrinsic correlation between vision and language itself could provide rich supervision signals. Considering its advantages, we take one step forward based on Video Swin Transformer that we add a language branch to view the whole framework as a multi-task model (VST+L) as shown in Figure~\ref{vl}. There are two alternatives for VST+L. The simple one only utilizes the action class name, e.g., \textit{spinning} as the language input; however, such a simple word could largely ignore important information related to the subject or the surrounding. Further, we propose to use detailed action descriptions as the textual input. Specifically, we first pre-define textural descriptions for the problem behavior classes by searching on the web and publications, as listed in Table~\ref{description}. During training, when the visual branch, i.e., Video Swin Transformer, takes as input a specific category of video (e.g., a ``head banging" action video), the corresponding action category description will be first tokenized into a language input tensor, and then the text encoder will encode the language input to a language feature. In this work,  we utilize the CLIP~\cite{radford2021learning} (Contrastive Language-Image Pretraining) text encoder as our text encoder. The original CLIP model consists of an image Transformer encoder and a text Transformer encoder and is trained on a large-scale image-text pairs dataset. The supervision contained in the natural language provides strong cross-modal representation ability for both CLIP encoders. In our work, in order to introduce additional supervision without bringing in extra annotations, we utilize the abundant image-text knowledge in the CLIP text encoder. As shown in Figure~\ref{vl}, since the language feature from the text encoder actually contains sufficient corresponding visual information, we propose to minimize the distance between the visual feature from the Video Swin Transformer and the language feature in the feature space for the input visual and text pair (i.e., input video and its corresponding action category description). With such a strategy, the additional language knowledge is actually distilled into the visual branch, introducing more information for the recognition task and making the language feature more predictable. We achieve such an operation by introducing a contrastive loss based on cosine similarity as follows:
\begin{equation}
L_{contrastive} = - \frac{v \cdot l}{|v|\cdot|l|},
\end{equation}
where $v, l$ is the trainable visual features and the fixed language features, respectively. For action classification training, we still use the cross-entropy loss $L_{CE}$ based on the ground truth action labels, then the total loss function can be formulated as:
\begin{equation}
L = L_{CE} + \lambda\cdot L_{contrastive},
\end{equation}
where $\lambda$ is a trade-off hyperparameter.
Based on this strategy, the visual feature can be pulled closer to the textual feature which corresponds to the same action category since it has been trained on a large-scale vision-language dataset, thereby improving the discriminative power of the visual feature. 
\textbf{We emphasize it here that the action descriptions we introduce do not add substantial annotation burden, since we generate the description for an action/behavior category, not for an individual video.}
It also should be noted that we only use language supervision (text branch) during model training; in the test/inference stage, we remove the text branch and only use the trained Video Swin Transformer to predict the action category of a test video. Therefore, our proposed method will not bring any extra computational cost in inference.
\section{Experiments}

\subsection{Experimental settings}
In this subsection, we introduce our training and evaluation settings. For a fair comparison, we follow previous works and utilize 5-fold cross-validation and report both top-1 accuracy and the F1 score.
During training, we first do regular data augmentation following the Video Swin Transformer training on Kinetics400~\cite{carreira2017quo} dataset, i.e., we resize the frames so that the shortest side is 256, and then we random crop a sub-region and resize it into $224 \times 224$. Additionally, we also flip the frame with a probability of 0.5 during training. We also normalize the frames into [0, 1]. During optimization, we select the most common optimizer AdamW~\cite{reddi2019convergence}. The initial learning rate is 0.001 and the weight decay is 0.05. We have two settings for the total training epochs. If the Video Swin Transformer backbone is pretrained on Kinetics400, we fine-tune another 50 epochs on the autism behavior dataset (this is denoted as \textbf{VST(pretrained)}); if the backbone is trained from scratch, the training epoch is set as 100 (this is denoted as \textbf{VST}). For both settings, we use the Cosine Annealing learning rate scheduler. We implement our method using Pytorch and carry out the experiments using 4 A5000 GPUs. 




\subsection{Experimental results}
\label{sec:result}
\textbf{Vision-only model}. We first evaluate the Video Swin Transformer on SSBD and ESBD datasets. The results are reported in Table~\ref{ESBD} and Table~\ref{SSBD}. It could be obviously observed from the results that Video Swin Transformer (\textbf{VST}) outperforms all the previous methods by a significant margin (i.e., the improvement is higher than 9$\%$ between our VST and the previous state-of-the-art HOF-BOVW method~\cite{negin2021vision} on ESBD dataset), which demonstrates its superior performance and versatility in different data domains. Moreover, we can also notice that using the pre-trained model on the large-scale general action dataset (Kinetics400~\cite{carreira2017quo}) can further improve the recognition performance, indicating the visual knowledge related to human actions obtained from Kinetics400 has good transferability on other specific action domains, such as autistic behaviors.   

\begin{table}[h]\centering
\caption{Comparison with previous methods on ESBD. VST means Video Swin Transformer 
and VST+L means adding our language supervision into VST ((d) means using descriptions as language input).}
\setlength{\tabcolsep}{3mm}{
\begin{tabular}{c|c|c}
\toprule[1pt]
Method              & Acc($\%$)  & F1 Score  \\ \hline
3DCNN~\cite{negin2021vision}               & 42   & $\_$  \\
ConvLSTM~\cite{negin2021vision}              & 74   & $\_$   \\
HOF-BOVW~\cite{negin2021vision}              & 79   & $\_$   \\
Skeleton-BOVW~\cite{negin2021vision}        &61    & $\_$    \\
Skeleton-LSTM~\cite{negin2021vision}        &59    & $\_$    \\\hline
VST          &      88.50  & 89.03   \\
VST(pretrained)  & \textbf{90.94}  & \textbf{91.58}  \\ \hline
VST+L(d)          &      90.04  &  93.33  \\
VST+L(d)(pretrained)  & \textbf{94.43} & \textbf{96.21}   \\ \bottomrule[1pt]
\end{tabular}}
\label{ESBD}
\end{table}

\begin{table}[h]\centering
\vspace{-1em}
\caption{Comparison with previous methods on SSBD. VST means Video Swin Transformer 
and VST+L means adding our language supervision into VST((d) means using descriptions as language input).}
\setlength{\tabcolsep}{3mm}{
\begin{tabular}{c|c|c}
\toprule[1pt]
Method              &  Top1 ($\%$) & F1 Score    \\ \hline
HDM~\cite{rajagopalan2014detecting}   & 73.6  & $\_$ \\
I3D(RGB)~\cite{ali2022video}               & 76.92  & 60   \\
I3D(RGB+Flow)~\cite{ali2022video}     & 75.62  & 69   \\
MS-TCN++(I3D)~\cite{wei2022vision}    &  $\_$  & 78   \\
MS-TCN(I3D)~\cite{wei2022vision}     &  $\_$ & 83   \\ \hline
VST          &      94.66  & 92.37  \\
VST(pretrained)  & \textbf{95.63} &  \textbf{95.76}  \\ \hline
VST+L(d)          &      96.25   & 96.18 \\
VST+L(d)(pretrained)  & \textbf{97.40}  & \textbf{97.27}  \\ \bottomrule[1pt]
\end{tabular}}
\label{SSBD}
\end{table}

\begin{figure}[t]
\centering
\subcaptionbox{ESBD}{\includegraphics[width=0.48\columnwidth]{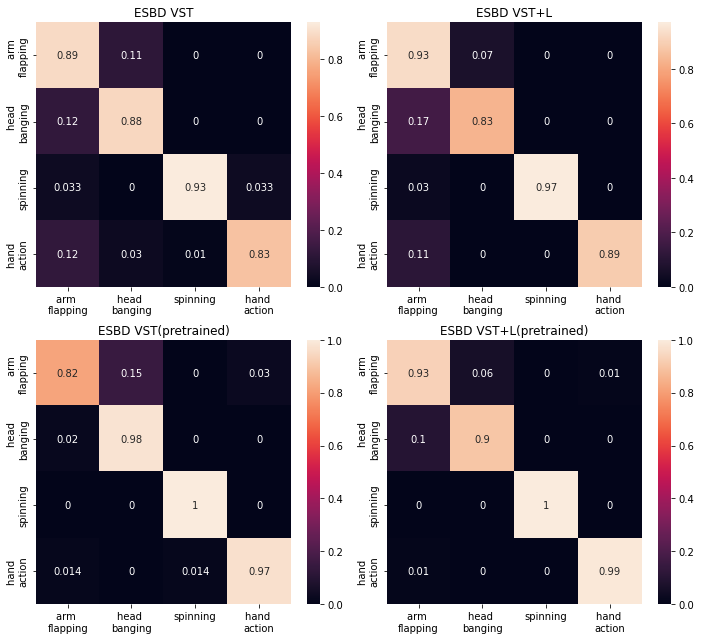}} 
\subcaptionbox{SSBD}{\includegraphics[width=0.48\columnwidth]{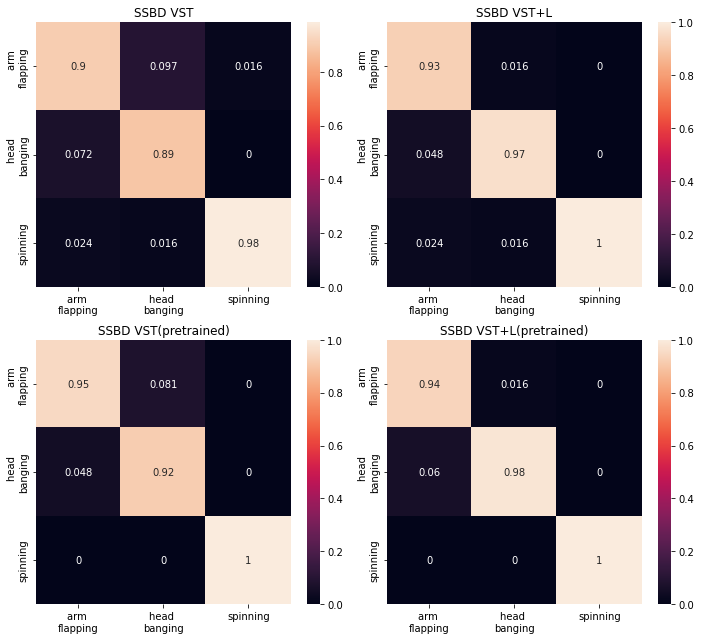}}
\caption{Illustration of the confusion matrix on (a)ESBD and (b)SSBD. "+L" means language supervision and "pretrained" means the visual branch uses the Kinetics400 pre-trained model and then is fine-tuned on ESBD or SSBD. It is obvious that both language supervision and pretraining could generally result in performance improvement.}
\label{cf_mat}
\end{figure}

\noindent\textbf{Vision-language model}. Further, we evaluate our vision-language model (VST+L) on ESBD and SSBD with the trade-off parameter $\lambda$ as 1 and 0.5, respectively. As shown in Table~\ref{ESBD} and ~\ref{SSBD}, the results suggest that language could indeed bring additional vision-related information via a well-trained language encoder. Especially for ESBD, compared with the results in Table~\ref{ESBD}, the language supervision brings a performance boost of 3.49$\%$ with a pretrained backbone. 
\textbf{It is noteworthy that, during inference, VST and VST+L share the same architecture, the only difference exists in the training stage that the visual network in VST+L learns additional visual-language knowledge from our action descriptions while the visual network in VST does not.}
In order to analyze the improvement in detail, we further illustrate the confusion matrix as shown in Figure~\ref{cf_mat}. For ESBD, as can be seen from the top row, when trained from scratch, introducing language supervision brings improvement for all classes. In the pretrained case, adding language supervision slightly harms the performance on \textit{head banging}, but largely boosts \textit{arm flapping} as well as the overall performance. For SSBD, for both pretraining and training from scratch, language brings consistent improvement. In addition, we observe that \textit{spinning} is much more distinguished than other classes in both ESBD and SSBD. 
Moreover, we found that using detailed descriptions of the behaviors shown in Table~\ref{text input} can obtain better performance than directly using the behavior name (e.g., spinning) as the textual embedding, which is intuitive since the detailed descriptions could provide more information than the behavior name.

\begin{table}[h]\centering
\caption{Comparison of top-1 accuracy when using different textual embedding inputs in our VST+L version. We found that it is more effective to use the detailed description from Table~\ref{description} than the category name of the ASD behaviors. }
\setlength{\tabcolsep}{3mm}{
\begin{tabular}{c|c|c}
\toprule[1pt]
Textual input        & VST+L(w)     & VST+L(d)   \\ \hline
ESDB           &    89.41           &  \textbf{90.04}  \\
SSDB           &    96.01           &  \textbf{96.25}  \\ \hline
\end{tabular}}
\label{text input}
\end{table}

\noindent\textbf{Joint dataset}. Moreover, to further validate the effectiveness of our method, we build a joint dataset by combining all the "Arm Flapping", "Head Banging", and "Spinning" videos from ESBD and SDBD datasets. In this way, the joint dataset is larger in terms of the sample size. Similarly,  we train and evaluate following the same protocol as the experiment of ESBD and SSBD (i.e., 5-fold cross-validation). As shown in Table~\ref{Joint}, introducing language supervision again brings consistent improvement (2.59$\%$ w/o pretrained visual branch and 1.92$\%$ w/ pretrained visual branch, respectively) in this setting. 

\begin{table}[h]\centering
\caption{Comparison with previous methods on the joint dataset. VST means Video Swin Transformer and VST+L(d) means adding our detailed language supervision into VST. ``w/ pretrained'' or ``w/o pretrained'' indicates the method is trained with or without the Kinetics400 pretrained model. }
\setlength{\tabcolsep}{3mm}{
\begin{tabular}{c|c|c}
\toprule[1pt]
Method        & w/o pretrained      & w/ pretrained     \\ \hline
VST           &    91.71           &  93.55  \\
VST+L(d)         &    \textbf{94.30}           & \textbf{95.47}   \\ \hline
\end{tabular}}
\label{Joint}
\end{table}

\begin{figure}[htp!]
\centering
    \includegraphics[width=12cm]{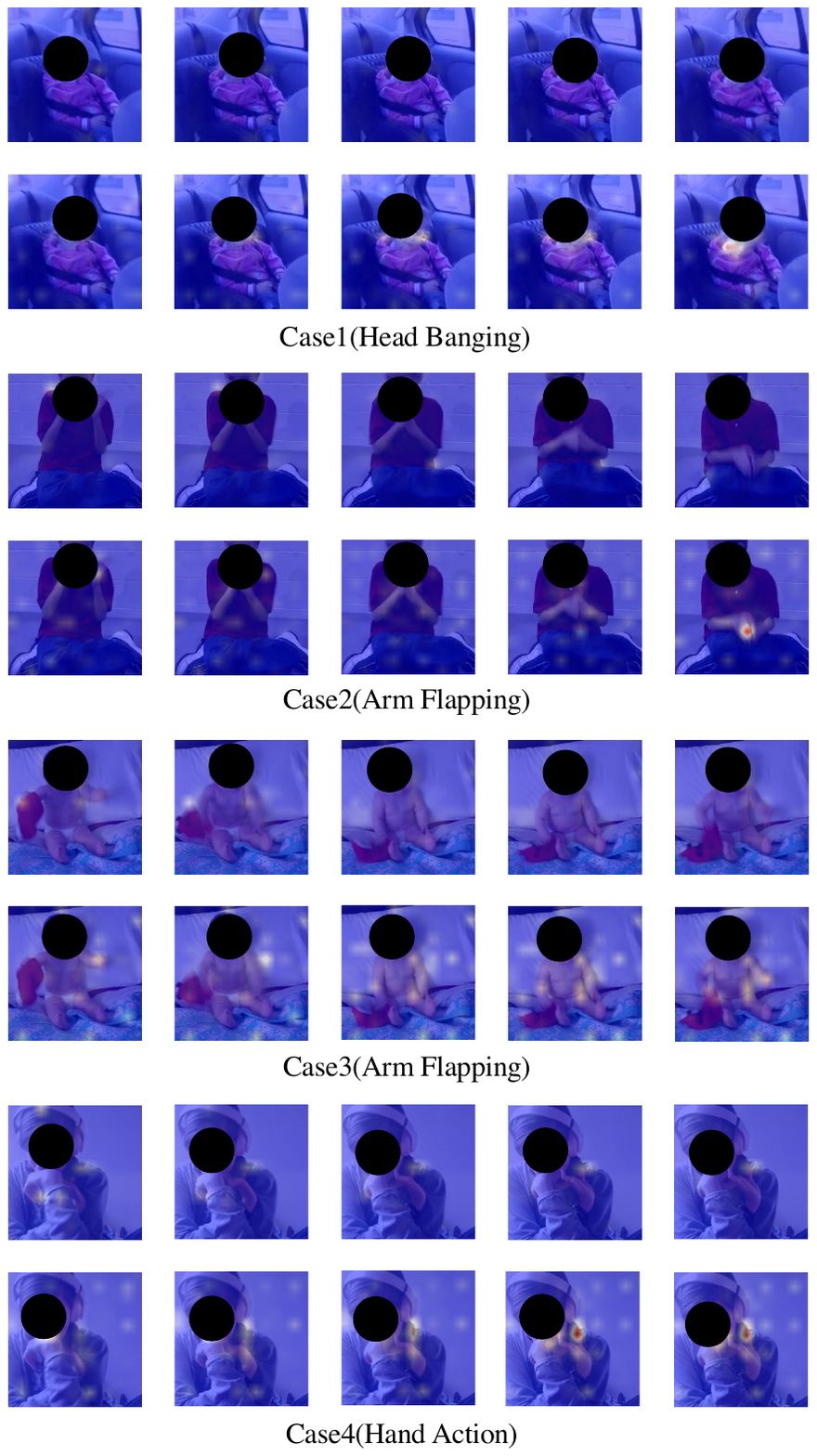}
    \vspace{-1.8em}
    \caption{Visualization of the attention map in the last attention layer of VST and VST+L. The first row of each case shows the attention map without language supervision (VST) and the second row shows the attention map with language supervision (VST+L).  Pixels with high attention values indicate regions that are most relevant to the final prediction. It is evident that using additional language supervision can make the model focus more on the relevant regions to the target action. } 
    \label{vis}
\end{figure}

\begin{figure}[t]
\centering
    \includegraphics[width=13cm]{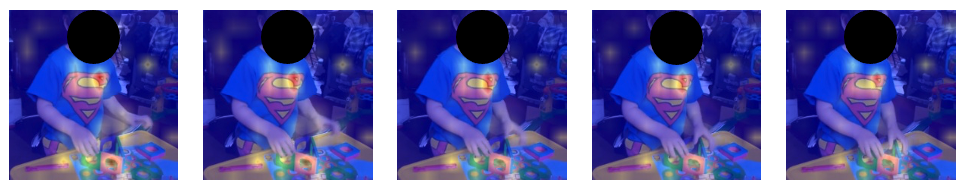}
    \caption{Visualization of a failure case in ESBD of the proposed language-assisted model for autism action recognition. It can be noticed that, in this 'Arm Flapping' case, the most attentive region appears in the logo of the T-shirt rather than the arms, leading to a false prediction of \textit{Spinning}.}
    \label{fail}
\end{figure}

\noindent\textbf{Visualizations}. The attention weights in the VST model or VST+L model contain semantic correlations between the pixels and the target action labels. Therefore, visualizing the attention weights can provide informative clues about how the model makes a prediction. We expect that our language-assisted model (i.e., VST+L) could further amplify the attention weights in regions that are most related to the action labels compared with the vanilla VST. Specifically, we visualize the first attention weights (there are two window attentions in each block) in the last attention block in the Video Swin Transformer (w/ and w/o our language supervision) to intuitively show the efficacy of our method. The shape of the attention weights is $B \times N \times N$, where $B$ is the batch size and $N=t \times h \times w$ is the multiplication of the spatiotemporal dimension in the current layer feature dimension. We first obtain its maximum along the second dimension as $B \times N$ and then reshape it into $B \times t \times h \times w$. To fit with the original input, we use trilinear interpolation to obtain a $B \times 32 \times 224 \times 224$ attention mask and we normalize it into [0, 1]. The final visualization is obtained by multiplying the attention masks by the original frames. 

As shown in Figure~\ref{vis}, introducing language information can indeed increase the attention weights of the region of interest and can show a tracking pattern towards the target region. For instance, in Case 1, it is obvious that using language supervision boosts higher attention weights on the head region, which is most related to the \textit{Head Banging} action. In Case 2, the model without language supervision fails to locate the child's hands, resulting in the false prediction of \textit{Spinning}, while the one with language supervision successfully distinguishes the regions that are highly related to the action. In Case 4, it is obvious that language supervision improves the intensity of the attention weights, which means that the model is more confident that the activated region relates to the target action class. 

However, our method also fails to recognize the correct actions in several videos. 
We provide a failure case with the ground truth label \textit{Arm Flapping} in Figure~\ref{fail}. Our method fails to precisely focus on the region of arms and generates wrong attention weights, leading to a false action prediction of \textit{Spinning}. Generally, the failure cases for our approach usually are those confusing or hard video samples, where the actions showcase weak visual patterns. For instance, in Figure~\ref{fail}, the kid playing with the toys does not show obvious ”Arm Flapping” behaviors, and the upper body is the most salient part in the frames, which makes the model focus on it and predict it as ”Spinning” because the most discriminative visual features for ”Spinning” is the bodies. Besides, weak visual patterns in video frames may also lead to misalignment between vision and language, which does harm to the knowledge distillation from the pre-trained CLIP text encoder. In future work, collecting a new dataset with a larger size could alleviate this issue because more training samples could neutralize the negative effect brought by the small portion of the noisy samples.

\section{Conclusion and Discussion}
Autism Spectrum Disorder (ASD) is a risky health problem for infants and toddlers. During the treatment, correctly distinguishing the stimming behavior is of critical importance in early diagnosis. In this work, we utilize one of the state-of-the-art action recognition models, i.e., Video Swin Transformer (VST), for this challenging task. The experimental results on two popular autistic action recognition datasets show that VST significantly outperforms the previous methods. Based on this strong baseline, we move forward to leverage the action description for each class and incorporate a pretrained language encoder to extract language features, pulling closer the distance between visual representation and textual one during training. By this language-assisted training strategy, the model can learn better visual representation for improving recognition performance. 
\textbf{The proposed method does not introduce additional annotation efforts, since the language description is shared for all videos in the same category. Meanwhile, the language network will be discarded during inference after the visual-language knowledge has been well learned in our visual network during training.} The final performance validates the effectiveness of using additional language supervision for autistic action recognition. 

Based on this work, it is clear that AI technology could largely help the development of ASD diagnosis, which indicates that, in the future, more possibilities should be in-depth investigated. For instance, during treatment, the physician usually provides an order to the autistic child to see the reaction, on which the diagnosis will be made. And it has been solid evidence that the intensity of the noise from the environment has a lot to do with the symptom~\cite{kanakri2017noise,landon2016qualitative}.
These suggest that signals from different modalities (e.g., vision and sound) could all be possible to facilitate accurate autistic behavior analysis. 
Also, the stimming behavior shows a strong repeated action pattern, and its frequency is also related to the patient's condition. Such periodic actions can be well-detected by modern computer vision algorithms. These examples can provide the potential possibility of the next step towards ASD studies combined with AI. Besides, a more effective dataset could be built for this task, since in ESBD and SSBD, all the data are autism behaviors. However, in real applications, we have to recognize stimming actions from  massive normal actions, which is more like an anomaly detection~\cite{chandola2009anomaly} manner. We will develop solutions to address these aforementioned potential challenges in our future work. 


 \bibliographystyle{elsarticle-num} 
 \bibliography{sample}





\end{document}